\documentclass[sigplan,nonacm]{acmart}
\usepackage{pifont}
\usepackage{enumitem}
\usepackage{graphicx}
\usepackage{subfigure}
\usepackage{multirow}
\usepackage{tablefootnote}
\usepackage{threeparttable}
\usepackage[vlined,linesnumbered,ruled]{algorithm2e}
\SetArgSty{textnormal} 
\usepackage{color}
\usepackage[noend]{algpseudocode}
\usepackage{xspace}
\usepackage{fancyhdr}
\usepackage{multicol}
\usepackage{amsfonts}
\usepackage{ulem}
\usepackage{tikz}
\usepackage{balance} 
\usepackage{makecell}
\usepackage{booktabs}

\def\ie{\textit{i.e.}\xspace}
\def\eg{\textit{e.g.}\xspace}

\def\method{\textsf{Apex}\xspace}

\newcommand*{\circled}[1]{\lower.8ex\hbox{\tikz\draw (0pt, 0pt)%
    circle (.47em) node {\makebox[0.4em][c]{\small #1}};}}

\begin{document}
 
\title[Apex]{Efficient CPU-GPU collaborative MoE inference via Adaptive Expert Assignment}

\author{Wenhui Lv, Yunming Liao, Hongli Xu, Yang Xu, Junhao Cheng}

\begin{abstract}
Mixture-of-Experts (MoE) models reduce inference computation through sparse expert activation, but their large expert memory footprint often exceeds GPU capacity, making expert offloading necessary for memory-constrained inference. However, existing systems typically handle GPU-missed experts by either transferring them to the GPU, incurring substantial PCIe transfer latency, or executing them on the CPU, making the slower CPU computation the bottleneck. To address these limitations, we present \method, a CPU-GPU collaborative MoE inference system that adaptively assigns these missed experts to the CPU or GPU for efficient execution. By exploiting differences in expert token-loads, \method selectively transfers high-load experts to the GPU while executing low-load experts on the CPU. To minimize inference latency, \method adaptively adjusts expert assignments to balance the CPU computation path against the GPU transfer-and-computation path, allowing both execution paths to complete as early as possible. Extensive experiments on three representative MoE LLMs show that \method improves prefill speed by up to 51.9\% and decode throughput by up to 39.5\% over the strongest baseline across the evaluated batch sizes, with consistent gains under different GPU-memory budgets.
\end{abstract} 


\maketitle

\section{Introduction}\label{sec_introduction}
Mixture-of-Experts (MoE)~\cite{deepseekai2025deepseekv3technicalreport,jiang2024mixtralexperts,openai2025gptoss120bgptoss20bmodel,glm5team2026glm5vibecodingagentic,kimiteam2026kimik3openfrontier} architectures have emerged as an effective way to scale the capacity of Large Language Models (LLMs)~\cite{3495724.3495883,zhang2022optopenpretrainedtransformer,geminiteam2025geminifamilyhighlycapable,touvron2023llama2openfoundation,openai2024gpt4technicalreport} without proportionally increasing inference computation. Instead of applying the same dense feed-forward network to every token, an MoE module contains many specialized expert networks and routes each token to only a small subset of them~\cite{Cai_2025}. 
Despite this computational advantage, MoE models still impose a substantial memory footprint~\cite{xue2025moeinfinityefficientmoeinference,tang2024hobbitmixedprecisionexpert}. 
Sparse activation cannot reduce storage requirements, as all expert parameters must remain accessible during inference~\cite{rajbhandari2022deepspeedmoeadvancingmixtureofexpertsinference}. 
Consequently, storing the entire model in limited and expensive GPU memory poses difficulties for private and on-premises deployment~\cite{10695719,chen2026bpdqbitplanedecompositionquantization,10.1145/3714983.3714987,3618408.3619696}. 
Expert offloading~\cite{10.1109/ISCA59077.2024.00078,10.1145/3676536.3676741,song2025promoefastmoebasedllm} provides a practical solution by storing expert weights in high-capacity CPU host memory. In this context, some activated experts are inevitably absent from GPU memory at runtime, referred to as \textit{GPU-missed experts}. A fundamental question for offloaded MoE inference is therefore how to process these missed experts. 

One option is to \textit{execute GPU-missed experts on the GPU}, where the missed experts must first be transferred from host memory over PCI Express (PCIe) before execution. 
However, GPU misses occur frequently, and each miss incurs an expensive expert transfer that is much slower than GPU execution. Even with half of the experts retained on the GPU, over 50\% of the required experts frequently miss GPU residency~\cite{yan2025acceleratingmixtureofexpertinferenceadaptive,eliseev2023fastinferencemixtureofexpertslanguage}. 
Prior work~\cite{tang2024hobbitmixedprecisionexpert} shows that on-demand expert transfers slow inference by more than $17\times$. 
Existing systems commonly employ speculative prefetching~\cite{eliseev2023fastinferencemixtureofexpertslanguage} to hide part of this cost. 
They predict the experts required by the next layer from the current layer's hidden states and transfer the predicted experts in parallel with ongoing computation. However, the available overlap is bounded by approximately one Transformer layer of computation. 
Our preliminary experiments in Sec.~\ref{sec:motivation} show that transferring the required experts takes at least 5.2$\times$ as long as GPU computation, leaving most of the transfer latency exposed even with prefetching. Pure GPU execution therefore remains transfer-bound.

The other option is to \textit{execute GPU-missed experts on the CPU}. 
Instead of transferring the expert weights, the GPU sends the much smaller expert inputs to the CPU, which computes the missed experts using weights already in host memory and returns their outputs to the GPU.
Meanwhile, the GPU computes its resident experts and then merges the outputs from both devices. 
This approach transfers only the much smaller inputs and outputs rather than entire experts. 
However, CPUs provide far lower expert-computation throughput than GPUs, commonly delivering more than 10$\times$ lower bfloat16 compute throughput~\cite{li2025largelanguagemodelinference,10538369,10.1007/978-3-032-07612-0_10}. 
The GPU must wait for CPU results at each layer, placing CPU execution on the critical path. This bottleneck becomes more severe at the larger batch sizes as the CPU processes a growing expert workload. 
Existing systems attempt to reduce GPU misses and thereby limit the workload assigned to the CPU. For example, they profile expert popularity offline and keep globally popular experts resident in GPU memory~\cite{kamahori2025fiddlercpugpuorchestrationfast}, or transfer GPU-missed experts after each inference step to improve subsequent GPU hit rates~\cite{11420335}.
Nevertheless, GPU memory remains limited, and the activated experts vary across inference requests, so the CPU inevitably processes part of the expert workload. Its much longer execution time (\eg, more than $5.4\times$ the GPU time in our preliminary study in Sec.~\ref{sec:motivation}) continues to slow down overall inference latency.

\begin{figure}[t]
    \centering
    \includegraphics[width=1.0\linewidth]{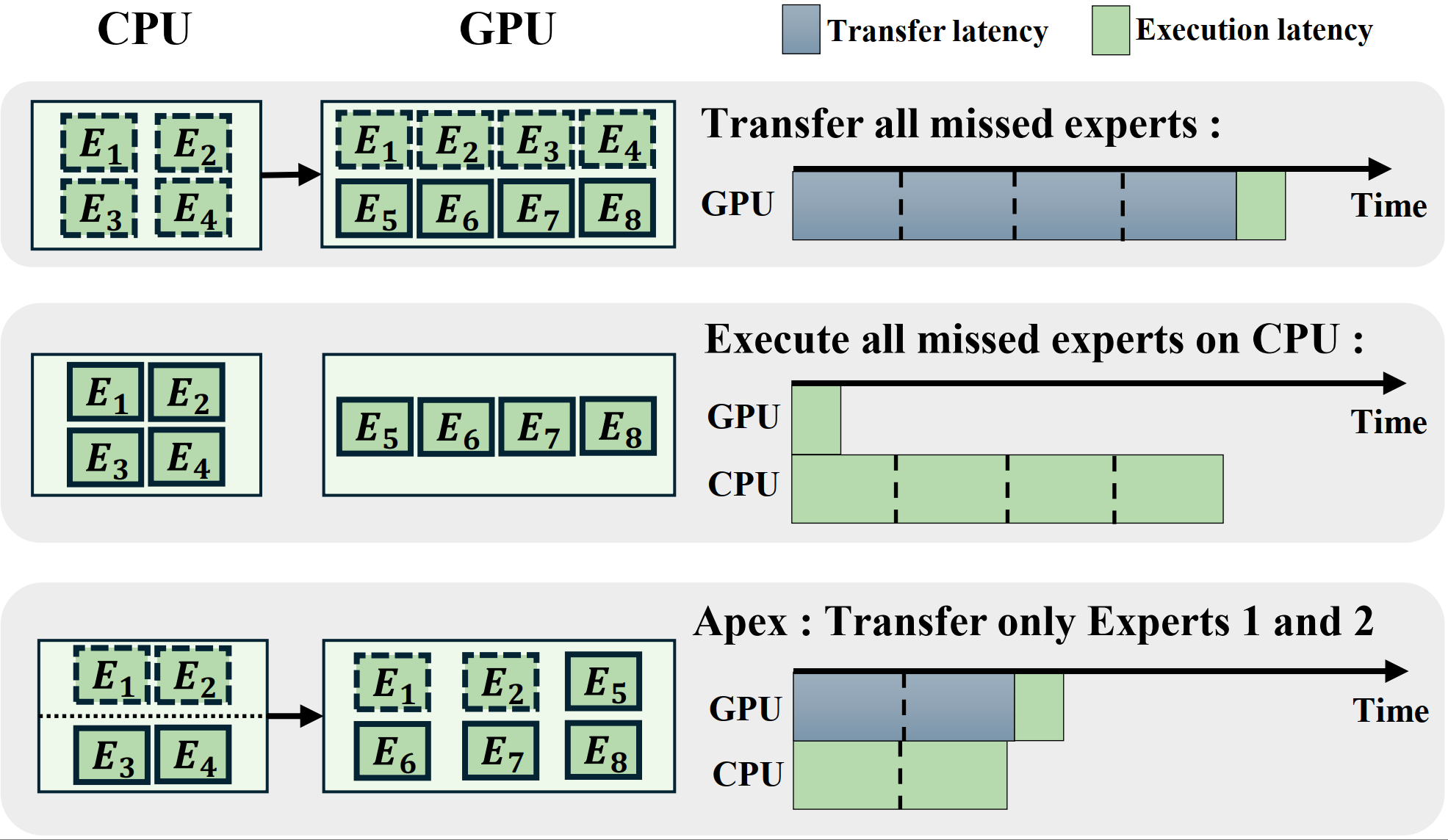}
    \vspace{-2mm}
    \caption{MoE stage latency under different inference paradigms. Dashed-outlined experts are transferred from the CPU to the GPU for execution.}\label{fig:case}
    \vspace{-3mm}
\end{figure}

We argue that the key challenge lies in deciding where to execute GPU-missed experts. Executing them on the GPU incurs transfer latency, whereas executing them on the CPU incurs computation latency.
To this end, we propose \method, an efficient CPU-GPU collaborative MoE inference system with adaptive expert assignment.
Rather than treating all GPU-missed experts uniformly, \method adaptively determines which experts should be transferred to the GPU for execution and which should remain on the CPU. 
We observe that tokens processed together in a serving batch are routed independently, producing highly variable numbers of tokens, \ie, \textit{token-loads}, across activated experts. An expert's token-load determines its computation cost, whereas the amount of expert weights that must be transferred remains unchanged. 
Consequently, transferring a high-load expert to the GPU is more worthwhile because the faster GPU computation offsets the transfer cost, whereas transferring a low-load expert introduces more transfer overhead than the computation time it saves.
\method exploits this insight by transferring high-load experts to the GPU for faster execution while directly executing low-load experts on the CPU to avoid unnecessary weight transfers. 
Through this adaptive execution assignment, \method substantially reduce the CPU workload while keeping GPU-side transfer latency under control. As illustrated in Figure~\ref{fig:case}, it avoids both excessive transfer latency and prolonged CPU computation, minimizing pipeline bubbles and reducing end-to-end inference latency.

However, realizing this adaptive assignment poses a significant challenge.
Under CPU-GPU collaborative execution, overall latency is determined by the slower of two concurrent paths. The CPU path consists of computing the experts assigned to the CPU, whereas the GPU path includes both transferring selected experts and executing all GPU-assigned experts. The number of transferred experts directly shapes both paths.
Transferring too many experts exposes excessive transfer latency and makes the GPU path the bottleneck. Transferring too few leaves the CPU with excessive computation, making the CPU path the bottleneck.
An effective assignment must therefore reduce enough CPU work to shorten the CPU path without introducing excessive transfer overhead on the GPU path. The goal is to minimize the maximum latency of the two concurrent paths. 
In summary, this paper makes the following contributions:
\begin{itemize}
    \item We propose \method, a CPU-GPU collaborative MoE inference system that adaptively assigns GPU-missed experts between CPU and GPU execution to balance the two paths and reduce inference latency.
    \item To minimize inference latency, \method combines speculative and exact assignment. Speculative assignment predicts high-load experts one layer ahead and starts their transfers early, extending the transfer window. Exact assignment uses an online, rate-driven algorithm to determine the remaining CPU/GPU assignments that minimize the layer latency. 
    \item We implement \method and evaluate it on three representative MoE models across different batch sizes and GPU memory budgets. Compared with the strongest baselines, Apex improves prefill speed by up to 51.9\% and decode throughput by up to 39.5\%, with average gains of 31.2\% and 31.1\%, respectively.
\end{itemize}

\section{Background and Motivation}\label{sec:prelim}
\subsection{MoE Inference}
Both dense and MoE-based LLMs stack $L$ Transformer layers with the same basic attention module~\cite{vaswani2023attentionneed}. Their key difference lies in the feed-forward network (FFN). A dense Transformer applies one FFN to every token, whereas an MoE Transformer replaces it with a Mixture-of-Experts module containing $E$ smaller FFNs, referred to as \textit{experts}~\cite{shazeer2017outrageouslylargeneuralnetworks,jiang2024mixtralexperts}. During inference, a router selects only the top-$K$ experts for each token, where $K\ll E$, and combines their outputs using the corresponding routing scores~\cite{lepikhin2020gshardscalinggiantmodels,dai2024deepseekmoeultimateexpertspecialization}. The sparse activation allows an MoE model to increase its capacity with sub-linear computational scaling~\cite{3586589.3586709,zoph2022stmoedesigningstabletransferable}. 

\textbf{Inference workflow.}
We first describe the forward pass of a single token. Let $\mathbf{x}_i$ denote its input hidden state at layer $i$. The attention module produces the state passed to the MoE router through a residual connection and normalization:
\begin{equation}
    \mathbf{h}_i
    =
    \operatorname{Norm}\!\left(
        \mathbf{x}_i+\operatorname{Attention}_i(\mathbf{x}_i)
    \right).
    \label{eq:attention_output}
\end{equation}
The router maps $\mathbf{h}_i$ to a normalized score vector $\mathbf{r}_i$ over all $E$ experts and selects the top-$K$ expert set $\mathcal{T}_i$:
\begin{equation}
    \mathbf{r}_i
    =
    \operatorname{softmax}\!\left(\operatorname{Router}_i(\mathbf{h}_i)\right),
    \qquad
    \mathcal{T}_i
    =
    \operatorname{TopK}(\mathbf{r}_i,K).
    \label{eq:moe_routing}
\end{equation}
For each selected expert $e\in\mathcal{T}_i$, we normalize its score within the selected set as
$\alpha_{i,e}=r_{i,e}/\sum_{j\in\mathcal{T}_i}r_{i,j}$. The selected experts process the token independently, and their outputs are combined according to these routing scores:
\begin{equation}
    \mathbf{m}_i
    =
    \sum_{e\in\mathcal{T}_i}
    \alpha_{i,e}\operatorname{Expert}_{i,e}(\mathbf{h}_i).
    \label{eq:moe_aggregation}
\end{equation}
Finally, the aggregated expert output is combined with $\mathbf{h}_i$ through another residual connection and normalization, producing
$\mathbf{x}_{i+1}=\operatorname{Norm}(\mathbf{h}_i+\mathbf{m}_i)$. This representation becomes the input to the next Transformer layer, and the same workflow repeats until the final layer completes.

\textbf{Expert routing under batching.}
LLM serving systems commonly use continuous batching~\cite{280922,10.1145/3600006.3613165} to improve hardware utilization and inference throughput. At each inference iteration, the system groups tokens from multiple requests into one batch~\cite{holmes2024deepspeedfastgenhighthroughputtextgeneration}. A prefill request contributes multiple prompt tokens, whereas each decoding request typically contributes one current token. Consequently, one model forward pass may process $N$ tokens contributed by different requests.

Let $\mathbf{X}_i=\{\mathbf{x}_{i,t}\}_{t=1}^{N}$ denote the input hidden states of these tokens at layer $i$, and let $\mathbf{H}_i=\{\mathbf{h}_{i,t}\}_{t=1}^{N}$ denote their post-attention states. Each token follows the workflow above and is routed independently to its own top-$K$ expert set $\mathcal{T}_{i,t}$. Thus, the batch produces $N\times K$ token-expert pairs in every MoE module. The set of experts activated by the entire batch is the union of the per-token selections:
\begin{equation}
    \mathcal{A}_i
    =
    \bigcup_{t=1}^{N}\mathcal{T}_{i,t}.
    \label{eq:batch_activated_experts}
\end{equation}
The token-loads are generally uneven across $\mathcal{A}_i$~\cite{gale2022megablocksefficientsparsetraining,hwang2023tuteladaptivemixtureofexpertsscale}. We define the \textit{token-load} of expert $e$ as the number of tokens routed to it:
\begin{equation}
    n_{i,e}
    =
    \sum_{t=1}^{N}\mathbb{I}\!\left(e\in\mathcal{T}_{i,t}\right).
    \label{eq:expert_load}
\end{equation}
Some activated experts may receive many tokens, whereas others receive only a few. This difference directly determines the amount of computation performed by each expert~\cite{hwang2023tuteladaptivemixtureofexpertsscale}.

\begin{figure}[t]
    \centering
    \includegraphics[width=1.0\linewidth]{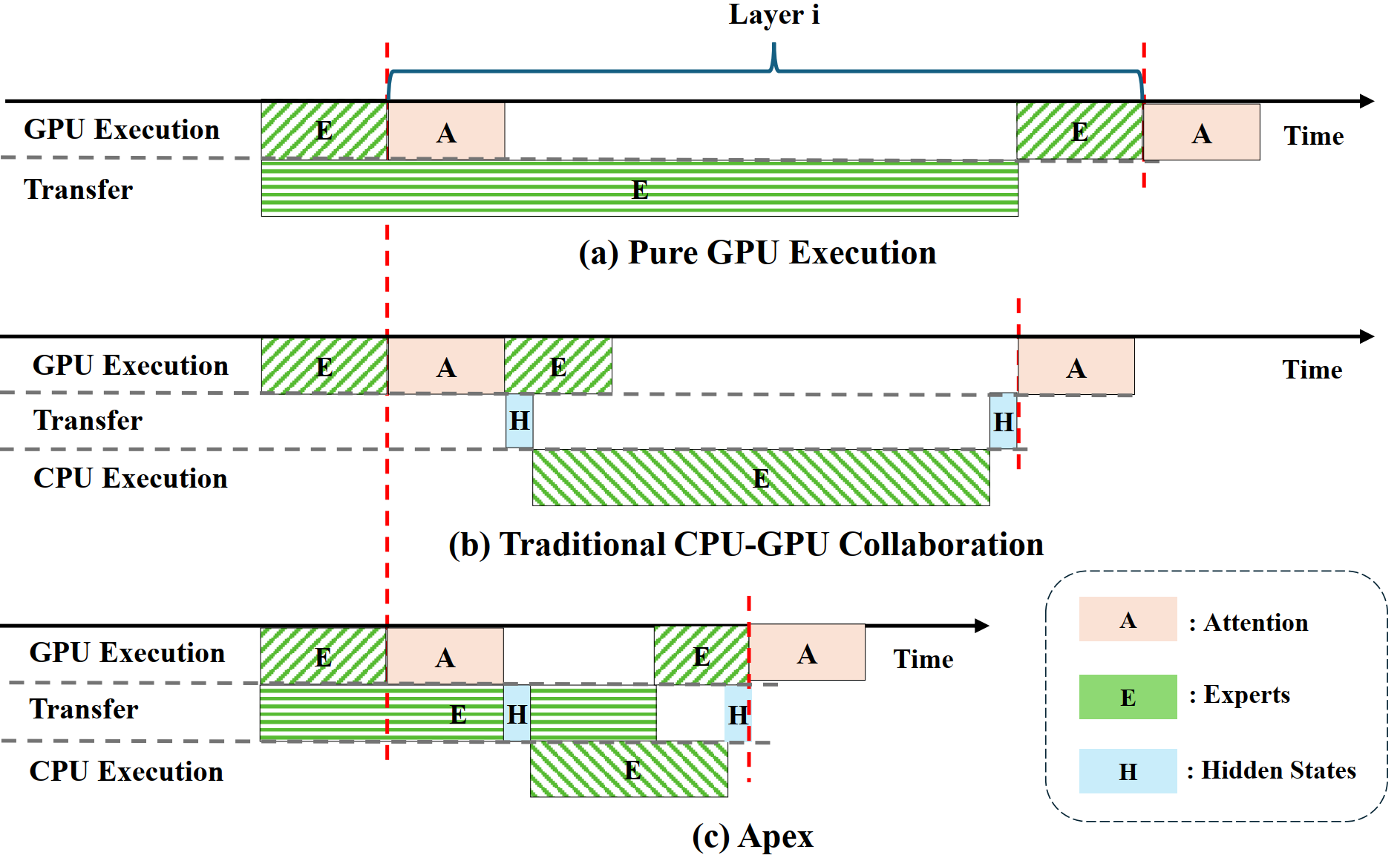}
    \caption{Execution timelines of pure GPU execution, traditional CPU-GPU collaboration, and \method.}\label{fig:timeline}
\end{figure}

\subsection{Execution with Expert Offloading}

Despite the computational efficiency, the large number of expert parameters imposes substantial memory requirements. 
Expert offloading addresses this issue by keeping the non-expert modules and only a limited subset of experts in GPU memory while keeping all experts in CPU host memory. 
When the router selects an expert that is absent from the GPU, existing systems generally follow one of two execution paradigms: (1) \textit{pure GPU execution} transfers every GPU-missed expert to the GPU and performs expert computation, whereas (2) \textit{CPU-GPU collaborative execution} computes GPU-missed experts in place on the CPU while the GPU processes its resident experts.

\textbf{Pure GPU execution.}
Pure GPU execution performs all expert computation on the GPU. An activated expert already resident in GPU memory can execute immediately, whereas the weights of a GPU-missed expert must traverse PCIe before execution. Existing systems use speculative prefetching to overlap transfers with ongoing computation and reduce the exposed transfer latency. For example, Mixtral-Offloading~\cite{eliseev2023fastinferencemixtureofexpertslanguage} estimates the experts required by the next layer from the current hidden states. For token $t$, it feeds hidden state $\mathbf{h}_{i,t}$ to the router of layer $i+1$ to obtain the predicted routing scores and expert set:
\begin{equation}
    \begin{aligned}
        \widehat{\mathbf{r}}_{i+1,t}
        &=\operatorname{softmax}\!\left(
            \operatorname{Router}_{i+1}(\mathbf{h}_{i,t})
        \right),\\
        \widehat{\mathcal{T}}_{i+1,t}
        &=\operatorname{TopK}\!\left(
            \widehat{\mathbf{r}}_{i+1,t},K
        \right).
    \end{aligned}
    \label{eq:prefetch_prediction}
\end{equation}
Although $\mathbf{h}_{i,t}$ is not the exact input to the next-layer router, residual connections keep the hidden states of adjacent layers highly similar~\cite{eliseev2023fastinferencemixtureofexpertslanguage,3691938.3691947}. The predicted expert set thus provides an early estimate of the next layer's routing result and can achieve considerable prediction accuracy (\eg, over 80\%~\cite{10.1145/3676536.3676741}). The system starts transferring these predicted experts while the current layer is still executing.

Speculative prefetching can hide only the transfers that fit within the computation window, which spans from the current layer's MoE module to the next layer's MoE module. Upon reaching the next layer's MoE module, any required expert that remains absent from the GPU must be transferred on demand. 
GPU expert computation is then blocked until the corresponding experts arrive, directly exposing PCIe latency on the GPU execution path.
Pure GPU execution therefore trades high GPU compute throughput for waiting on expert transfer, as illustrated in Figure~\ref{fig:timeline}(a).

\begin{figure*}[!t]
    \begingroup
    \setlength{\subfigcapskip}{1pt}
    \setlength{\subfigbottomskip}{0pt}
    \setlength{\abovecaptionskip}{3pt}
    \centering

    \makebox[\textwidth][c]{
        \subfigure[Batched expert activation
        \label{fig:motivation_a}]{
            \includegraphics[width=0.235\textwidth]
            {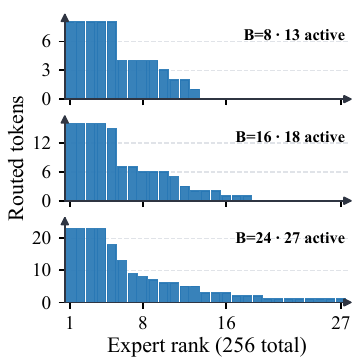}
        }
        \hspace{0.006\textwidth}
        \subfigure[Computation vs transfer
        \label{fig:motivation_b}]{
            \includegraphics[width=0.235\textwidth]
            {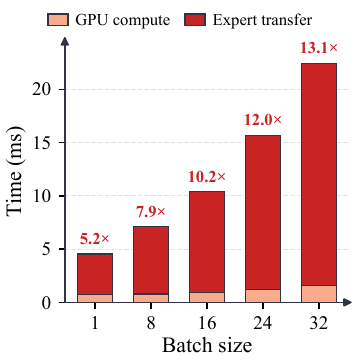}
        }
        \hspace{0.006\textwidth}
        \subfigure[CPU/GPU execution latency
        \label{fig:motivation_c}]{
            \includegraphics[width=0.235\textwidth]
            {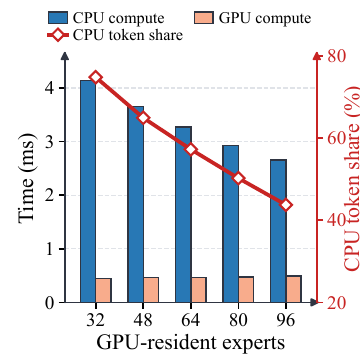}
        }
        \hspace{0.006\textwidth}
        \subfigure[CPU-GPU path balancing
        \label{fig:motivation_d}]{
            \includegraphics[width=0.235\textwidth]
            {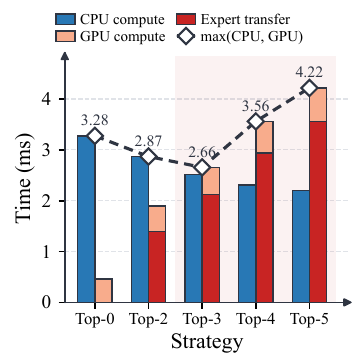}
        }
    }

    \caption{Preliminary experimental results.
    (a) Larger batches activate more experts with uneven token loads.
    (b) Transferring all GPU-missed experts dominates GPU computation.
    (c) Collaborative execution leaves inference CPU-bound.
    (d) Selectively transferring high-load experts reduces the CPU path,
    whereas excessive transfers shift the bottleneck to the GPU path.}
    \label{fig:moti-example}
    \endgroup
\end{figure*}

\textbf{CPU-GPU collaborative execution.}
CPU-GPU collaborative execution avoids transferring GPU-missed experts by evaluating them directly on the CPU. 
To reduce CPU computation overhead, existing works attempt to decrease the number of GPU misses and thereby limit the expert workload assigned to the CPU. 
A representative system, Fiddler~\cite{kamahori2025fiddlercpugpuorchestrationfast}, first profiles expert popularity on a calibration dataset and keeps the most frequently activated experts resident in GPU memory. Let $\mathcal{R}_i$ denote the resident expert set of layer $i$ under this residency policy. Once the router produces the exact token-expert assignments, the GPU groups the hidden states for each activated expert. For a GPU-missed expert $e\in\mathcal{A}_i\setminus\mathcal{R}_i$, its input group
$\mathbf{H}_{i,e}=\{\mathbf{h}_{i,t}\mid e\in\mathcal{T}_{i,t}\}$ is transferred to the CPU. The CPU evaluates these tokens using the experts in host memory, while the GPU concurrently evaluates the activated experts in $\mathcal{A}_i\cap\mathcal{R}_i$.

After CPU computation completes, its expert outputs are returned to the GPU. The GPU and CPU results are then combined using the original routing scores:
\begin{equation}
    \begin{aligned}
        \mathbf{m}^{\mathrm{GPU}}_{i,t}
        &=\sum_{e\in\mathcal{T}_{i,t}\cap\mathcal{R}_i}
        \alpha_{i,t,e}\operatorname{Expert}_{i,e}(\mathbf{h}_{i,t}),\\
        \mathbf{m}^{\mathrm{CPU}}_{i,t}
        &=\sum_{e\in\mathcal{T}_{i,t}\setminus\mathcal{R}_i}
        \alpha_{i,t,e}\operatorname{Expert}_{i,e}(\mathbf{h}_{i,t}),\\
        \mathbf{m}_{i,t}
        &=\mathbf{m}^{\mathrm{GPU}}_{i,t}
        +\mathbf{m}^{\mathrm{CPU}}_{i,t}.
    \end{aligned}
    \label{eq:cpu_gpu_collaborative_execution}
\end{equation}
Compared with transferring expert weights, transferring expert inputs and outputs introduces less PCIe traffic. Nevertheless, expert computation is still substantially slower on the CPU, as illustrated in Figure~\ref{fig:timeline}(b).

\subsection{Motivation for System Design}\label{sec:motivation}

To quantify the limitations of existing execution paradigms, we conduct preliminary experiments using Qwen3.5-35B-A3B~\cite{qwen3.5} on the MMLU~\cite{hendrycks2021measuringmassivemultitasklanguage} dataset. This model contains 256 experts in each MoE module and activates eight experts for every token. We combine multiple MMLU requests into batches and evaluate the resulting token-loads, transfer latency, and CPU/GPU execution latency. These experiments yield the following observations.

\textbf{Batched routing activates more experts with highly uneven token-loads.}
As shown in Figure~\ref{fig:motivation_a}, increasing the batch size from 8 to 16 and 24 expands the number of activated experts from 13 to 18 and 27, respectively. A larger activated set creates more potential GPU misses under a fixed GPU-memory budget. More importantly, token assignments remain highly uneven within each batch. At batch size 24, for example, the token-loads of the 27 activated experts range from one token to 23 tokens. Consequently, these experts impose substantially different computation workloads.

\textbf{Expert transfer dominates GPU computation.}
Under a fixed GPU-memory budget, larger activated sets generally create more GPU misses and hence more transfers under pure GPU execution.
Figure~\ref{fig:motivation_b} shows that, at batch size 1, expert transfer takes 3.80 ms per layer, compared with only 0.73 ms for GPU expert computation.
At batch size 32, expert transfer increases to 20.86 ms while GPU computation takes 1.59 ms, widening the ratio to $13.1\times$. Because speculative prefetching provides only a limited overlap window, a substantial fraction of this transfer remains exposed. Transferring every GPU-missed expert consequently makes pure GPU execution transfer-bound.

\textbf{Collaborative execution leaves inference CPU-bound.}
Following Fiddler~\cite{kamahori2025fiddlercpugpuorchestrationfast}, we profile expert popularity on the MMLU training split, keep the most popular experts resident on the GPU, and evaluate inference on the validation split at batch size 16. We vary the number of GPU-resident experts from 32 to 96. As shown in Figure~\ref{fig:motivation_c}, increasing the GPU-memory budget moves more computation to the GPU and gradually reduces CPU computation. With 32 experts resident on the GPU, the CPU still processes 74.80\% of the assignments and takes 3.66 ms per layer, compared with 0.45 ms on the GPU. Even with 96 GPU-resident experts, the CPU receives 43.73\% of the assignments and takes 2.98 ms, whereas GPU computation takes only 0.49 ms. Thus, CPU expert execution remains the long-tail latency bottleneck. 

These observations expose a common limitation of existing approaches: they treat all GPU-missed experts uniformly. Pure GPU execution transfers every missed expert, whereas traditional CPU-GPU collaborative execution computes every missed expert on the CPU. However, experts have identical transfer sizes while their token-loads differ substantially. Transferring a high-load expert can eliminate considerable CPU computation, whereas transferring a low-load expert incurs the same PCIe cost for little benefit. This insight motivates load-aware execution assignment: high-load experts should be transferred to the GPU, while low-load experts should remain on the CPU.

\textbf{Balancing the CPU and GPU execution paths.}
We next evaluate runtime, load-aware expert assignment with 64 experts resident on the GPU. Once the routing result of layer $i$ becomes available, we identify the GPU-missed expert set $\mathcal{M}_i=\mathcal{A}_i\setminus\mathcal{R}_i$ and rank its experts by their actual token-loads. The top-$P$ highest-load experts are assigned to GPU execution and their weights are transferred to the GPU, while all remaining GPU-missed experts execute on the CPU. Let $\mathcal{G}_i$ and $\mathcal{C}_i=\mathcal{M}_i\setminus\mathcal{G}_i$ denote these two sets, respectively.
Every expert in $\mathcal{G}_i$ contributes transfer latency to the GPU path, and the GPU waits for these transfers before performing expert computation. The layer latency for a given $P$ is determined by the longer of the CPU and GPU paths:
\begin{equation}
    T_i(P)
    =
    \max\!\left\{
        T_{\mathrm{CPU}}\!\left(\mathcal{C}_i\right),
        T_{\mathrm{transfer}}\!\left(\mathcal{G}_i\right)
        +T_{\mathrm{GPU}}\!\left(\mathcal{R}_i\cup\mathcal{G}_i\right)
    \right\}.
    \label{eq:layer_critical_path}
\end{equation}
Increasing $P$ moves the most computation-intensive GPU-missed experts from the CPU to the GPU, thereby shortening the CPU path. At the same time, each additional expert incurs a complete transfer latency, lengthening the GPU path. 

Figure~\ref{fig:motivation_d} illustrates this tradeoff. When $P=0$, the CPU path takes 3.28 ms, compared with only 0.46 ms for the GPU path, making CPU computation the bottleneck. Transferring the two highest-load experts introduces 1.39 ms of transfer latency but reduces the CPU path to 2.87 ms. The GPU path takes 1.89 ms, so CPU computation still determines layer latency. At $P=4$, the CPU path decreases to 2.52 ms, while the GPU path increases to 2.66 ms, including 2.12 ms of transfer latency. Although the GPU path becomes slightly longer, this setting achieves the lowest latency among the evaluated transfer counts. Increasing $P$ to 8 and 12 further reduces the CPU path to 2.31 ms and 2.21 ms, respectively, but raises the GPU path to 3.56 ms and 4.22 ms. Thus, transferring too many experts shifts the bottleneck from CPU computation to GPU-side transfer and increases overall latency.

These results motivate an adaptive assignment policy for \method, as shown in Figure~\ref{fig:timeline}(c).
At each layer of every inference step, \method uses the actual routing demand, GPU residency, and expert token-loads to determine which GPU-missed experts should execute on each device. The objective is to balance the CPU computation path against the GPU transfer-and-computation path, minimizing the longer of the two rather than applying a fixed transfer budget. Furthermore, substantial transfer latency limits the available transfer count before the GPU becomes the bottleneck. To provide greater flexibility for adaptive assignment, \method predicts high-load experts for the next layer and begins transferring during the current-layer. This speculative transfer creates an overlap window in which additional GPU-missed experts can be transferred without exposing transfer latency, allowing more CPU work to be reassigned to the GPU.

\section{System Design}\label{sec:design}
\subsection{Overview of \method} \label{sec:overview}

\begin{figure}[t]\centering
    \includegraphics[width=0.45\textwidth]{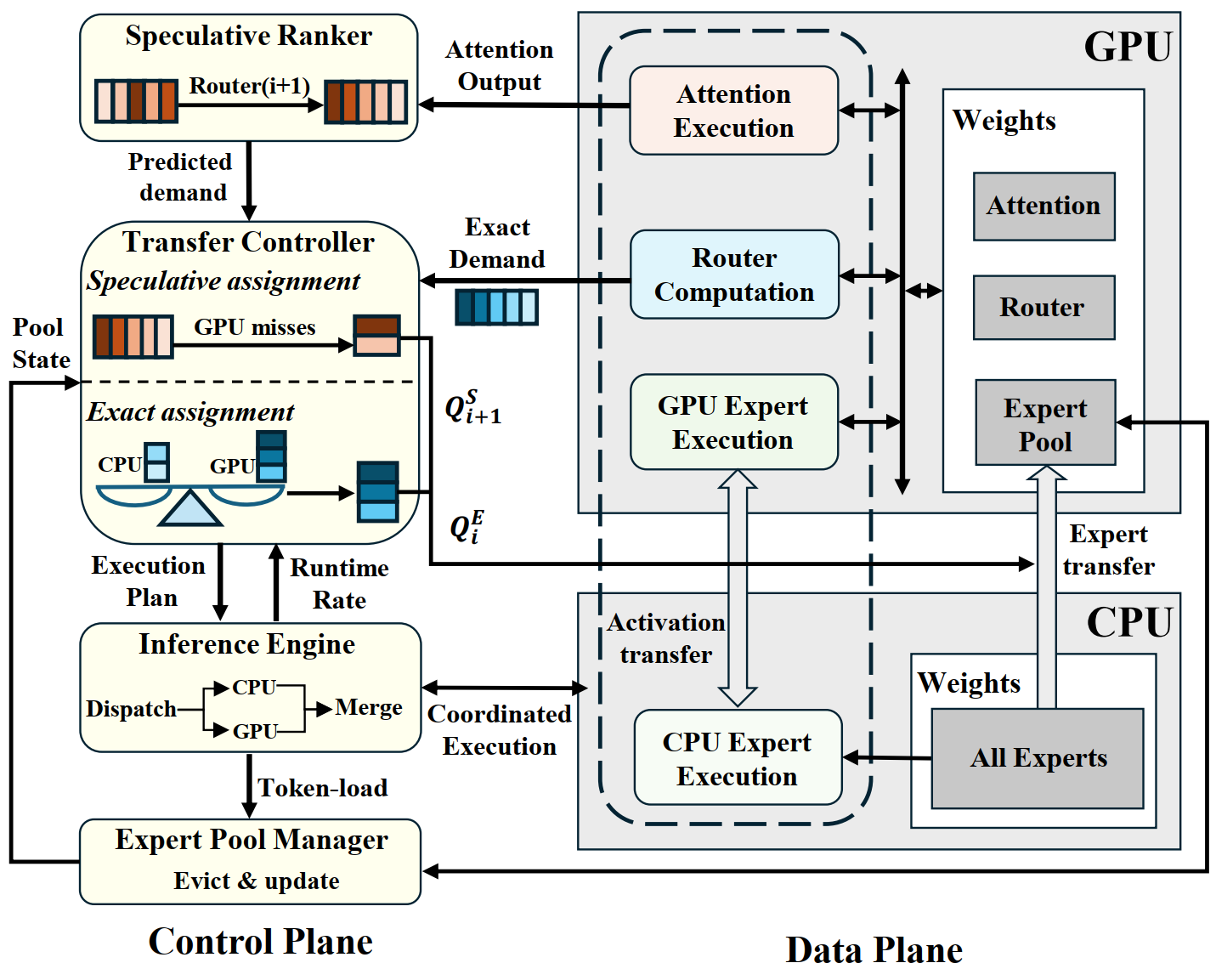}
    \caption{System overview of \method.} \label{fig:system_overview}
 \end{figure}

\method is an efficient MoE inference system with expert offloading, where all expert weights are stored in CPU host memory, while GPU memory holds the non-expert parameters and a fixed-capacity expert pool. 
Figure~\ref{fig:system_overview} illustrates how \method coordinates expert execution and transfers along the inference path. 
At each Transformer layer, inference proceeds from attention execution on GPU to the MoE block, which consists of router computation and CPU-GPU collaborative expert execution.
After attention completes, its output is passed to the router, which determines the exact expert demand and sends it to the \textit{transfer controller}. Using the current GPU pool state and observed runtime rates, the controller performs \textit{exact assignment}: it decides which GPU-missed experts should execute on CPU and which should be transferred to GPU, forming an expert execution plan and starting the exact expert transfers. The \textit{inference engine} follows this plan to group input tokens by expert and dispatch them to the assigned devices. CPU-assigned experts process activations transferred from GPU, while GPU-assigned experts execute after the exact expert transfers complete. The engine then returns CPU results to GPU and merges the expert outputs before proceeding to the next layer. 
Alongside this main inference path, the attention output of layer $i$ is also passed to the \textit{speculative ranker}, which applies the next layer's router to estimate expert demand for layer $i+1$. The ranker sends this prediction to the transfer controller for \textit{speculative assignment}. After current-layer exact transfers complete, the controller prefetches predicted experts, assigning some next-layer missed experts to GPU in advance while overlapping their transfers with ongoing computation. 
Throughout this process, the \textit{expert pool manager} maintains the GPU expert pool, selecting an eviction victim whenever an incoming expert requires a slot and reporting the updated pool state to the controller. The inference engine reports observed expert token-loads to the pool manager, which updates token-load-based priorities for every expert to make future eviction decisions.

\subsection{Speculative Ranker} \label{sec:expert_ranker}
At every layer, the speculative ranker estimates the experts that will be activated by the next layer, thereby exposing an early transfer window. Once the attention module of layer $i$ produces $\mathbf{H}_i=\{\mathbf{h}_{i,t}\}_{t=1}^{N}$, the speculative ranker feeds these representations to the router of layer $i+1$ and applies the prediction rule in Eq.~\eqref{eq:prefetch_prediction}. This prediction is used only to schedule prefetching and does not change the exact routing result or model output.

After obtaining the predicted token-level assignments, the ranker aggregates them into two statistics for each expert:
\begin{equation}
    \widehat{n}_{i+1,e}
    =
    \sum_{t=1}^{N}\mathbb{I}\left(e\in\widehat{\mathcal{T}}_{i+1,t}\right),
    \qquad
    \widehat{q}_{i+1,e}
    =
    \sum_{t:e\in\widehat{\mathcal{T}}_{i+1,t}}
    \widehat{r}_{i+1,t,e}.
    \label{eq:predicted_expert_statistics}
\end{equation}
Here, $\widehat{n}_{i+1,e}$ is the predicted token-load assigned to expert $e$, while $\widehat{q}_{i+1,e}$ is the sum of their predicted routing scores. Because expert computation is primarily determined by the number of routed tokens, the ranker orders the experts by decreasing $\widehat{n}_{i+1,e}$. When two experts have the same predicted load, the expert with the larger $\widehat{q}_{i+1,e}$ is placed earlier.

Let $\widehat{\mathcal{A}}_{i+1}=\{e\mid\widehat{n}_{i+1,e}>0\}$ denote the set of experts predicted to be active. At inference step $s$, the speculative ranker produces the ordered descriptor: 
\begin{equation}
    \widehat{\mathcal{D}}_{i+1}^{(s)}
    =
    \left[
    (i+1,\widehat{e}_r,\widehat{n}_{i+1,\widehat{e}_r},
    \widehat{q}_{i+1,\widehat{e}_r})
    \right]_{r=1}^{|\widehat{\mathcal{A}}_{i+1}|},
    \label{eq:predicted_descriptor}
\end{equation}
where $(\widehat{e}_1,\ldots,\widehat{e}_{|\widehat{\mathcal{A}}_{i+1}|})$ follows the ranking above. The ranker passes this descriptor to the transfer controller as the prefetch priority. The controller therefore transfers high-load experts first and overlaps their transfers with current computation.

\subsection{Transfer Controller} \label{sec:expert_placement}

The transfer controller combines \textit{exact expert assignment} for the current layer with \textit{speculative expert assignment} for the next layer. Together, they adapt CPU-GPU expert assignment to observed and predicted demand: exact assignment balances the current execution paths, while speculative assignment uses the available overlap window to place next-layer experts on GPU in advance. Exact transfers take priority over speculative transfers.

\textbf{Exact expert assignment.}
Let $\mathcal{A}_i$ denote the exact activated set of layer $i$, and let $\mathcal{R}_i$ denote the GPU resident set. The GPU-missed expert set is $\mathcal{M}_i=\mathcal{A}_i\setminus\mathcal{R}_i$.
The activated resident experts in $\mathcal{A}_i\cap\mathcal{R}_i$ execute directly on GPU. Exact expert assignment therefore focuses on the GPU-missed experts: it determines which of them should be transferred to GPU for execution and which should remain on CPU. 

Because transferring high-load experts removes more computation from CPU at the same transfer cost, the controller orders the missed experts as $(e_1,\ldots,e_m)$ so that
\begin{equation}
    n_{i,e_1}\ge n_{i,e_2}\ge\cdots\ge n_{i,e_m}.
    \label{eq:exact_load_order}
\end{equation}
For every transfer count $p$, the controller assigns the $p$ highest-load missed experts to GPU and leaves the remaining missed experts on CPU:
\begin{equation}
    \mathcal{G}_i(p)=\{e_1,\ldots,e_p\},
    \qquad
    \mathcal{C}_i(p)=\mathcal{M}_i\setminus\mathcal{G}_i(p).
    \label{eq:expert_partition}
\end{equation}
The transfer count $p$ determines the length of both concurrent execution paths. The CPU path contains the computation of the experts in $\mathcal{C}_i(p)$, while the GPU path contains the exposed transfer of the experts in $\mathcal{G}_i(p)$ followed by the computation of both resident and newly transferred experts. The layer latency is determined by the larger of the two. The assignment problem therefore reduces to selecting $p$ that minimizes this maximum.

To determine $p$, the controller performs online rate-driven balancing. It tracks three end-to-end service rates: CPU token throughput $v^{\mathrm{C}}$, GPU token throughput $v^{\mathrm{G}}$, and expert-transfer throughput $v^{\mathrm{T}}$. 
The first two rates measure the token processing throughput of the CPU and GPU during expert execution, while the third measures the expert transfer throughput.
The controller updates these rates from completed operations using an exponential moving average:
\begin{equation}
    v^{x,(s)}=(1-\beta)v^{x,(s-1)}
    +\beta\frac{w_x^{(s)}}{t_x^{(s)}},
    \qquad x\in\{\mathrm{C},\mathrm{G},\mathrm{T}\}.
    \label{eq:online_rate_update}
\end{equation}
For each possible transfer count $p$, the controller estimates the two concurrent execution paths as: 
\begin{equation}
    \begin{aligned}
        \widehat{T}^{\mathrm{C}}_i(p)
        &=\frac{\sum_{e\in\mathcal{C}_i(p)}n_{i,e}}{v^{\mathrm{C}}},\\
        \widehat{T}^{\mathrm{G}}_i(p)
        &=\frac{p}{v^{\mathrm{T}}}
        +\frac{\sum_{e\in(\mathcal{A}_i\cap\mathcal{R}_i)\cup\mathcal{G}_i(p)}n_{i,e}}{v^{\mathrm{G}}}.
    \end{aligned}
    \label{eq:estimated_execution_paths}
\end{equation}
The CPU estimate contains the computation assigned to CPU, whereas the GPU estimate contains the transfers followed by the computation of resident and newly transferred experts. The controller then selects $p^*$: 
\begin{equation}
    p_i^*
    =
    \arg\min_{0\le p\le m}
    \max\!\left\{
        \widehat{T}^{\mathrm{C}}_i(p),
        \widehat{T}^{\mathrm{G}}_i(p)
    \right\}.
    \label{eq:placement_objective}
\end{equation}
As $p$ increases, CPU work decreases, while transfer and GPU work increase. Scanning all possible transfer counts therefore directly identifies the approximate balance between the two execution paths under the latest observed rates. 
The controller assigns $\mathcal{C}_i(p_i^*)$ to CPU and $(\mathcal{A}_i\cap\mathcal{R}_i)\cup\mathcal{G}_i(p_i^*)$ to GPU. This partition forms the execution plan $\Pi_i^{(s)}$ passed to the inference engine.

After selecting $p_i^*$, the controller places the GPU-assigned misses in the exact transfer queue:
\begin{equation}
    \mathcal{Q}^{\mathrm{E}}_i=(e_1,\ldots,e_{p_i^*}).
    \label{eq:exact_transfer_queue}
\end{equation}
It immediately drains this queue to load the selected expert weights into the GPU pool, while the inference engine starts the CPU-assigned experts. Once the exact transfers complete, the engine executes the GPU-assigned experts, and the transfer channel becomes available for speculative assignment.

\begin{algorithm}[t]
\LinesNumbered
\caption{Adaptive assignment with exact and speculative transfers}
\label{alg:transfer_controller}
\KwIn{Activated set $\mathcal{A}_i$, loads $\{n_{i,e}\}$, resident sets $\mathcal{R}_i,\mathcal{R}_{i+1}$, prediction $\widehat{\mathcal{D}}_{i+1}^{(s)}$, rates $v^{\mathrm{C}},v^{\mathrm{G}},v^{\mathrm{T}}$.}
\KwOut{Execution plan $\Pi_i^{(s)}$ and transfer queues $\mathcal{Q}^{\mathrm{E}}_i,\mathcal{Q}^{\mathrm{S}}_{i+1}$.}
Identify GPU misses: $\mathcal{M}_i\gets\mathcal{A}_i\setminus\mathcal{R}_i$; $m\gets|\mathcal{M}_i|$\nllabel{line:tc_prepare_start}\;
Sort misses by token-load: $n_{i,e_1}\ge\cdots\ge n_{i,e_m}$\;
Initialize prefix load: $S_0\gets0$\;
\For{$p\gets1$ \KwTo $m$}{
    Accumulate token-load: $S_p\gets S_{p-1}+n_{i,e_p}$\;
}
Compute resident load: $L_{\mathrm{R}}\gets\sum_{e\in\mathcal{A}_i\cap\mathcal{R}_i}n_{i,e}$\nllabel{line:tc_prepare_end}\;
Initialize best choice: $T^*\gets\infty$; $p_i^*\gets0$\nllabel{line:tc_search_start}\;
\For{$p\gets0$ \KwTo $m$}{
    Estimate CPU latency: $\widehat{T}^{\mathrm{C}}_i(p)\gets(S_m-S_p)/v^{\mathrm{C}}$\;
    Estimate GPU path latency: $\widehat{T}^{\mathrm{G}}_i(p)\gets p/v^{\mathrm{T}}+(L_{\mathrm{R}}+S_p)/v^{\mathrm{G}}$\;
    Evaluate layer latency: $T_p\gets\max\{\widehat{T}^{\mathrm{C}}_i(p),\widehat{T}^{\mathrm{G}}_i(p)\}$\;
    \If{$T_p<T^*$}{
        Update best choice: $T^*\gets T_p$; $p_i^*\gets p$\nllabel{line:tc_search_end}\;
    }
}
Partition missed experts: $\mathcal{G}_i\gets\{e_r\mid1\le r\le p_i^*\}$; $\mathcal{C}_i\gets\mathcal{M}_i\setminus\mathcal{G}_i$\nllabel{line:tc_plan_start}\;
Build execution plan: $\Pi_i^{(s)}\gets\{\mathrm{CPU}:\mathcal{C}_i,\;\mathrm{GPU}:(\mathcal{A}_i\cap\mathcal{R}_i)\cup\mathcal{G}_i\}$\;
Build exact queue: $\mathcal{Q}^{\mathrm{E}}_i\gets(e_r)_{r=1}^{p_i^*}$\nllabel{line:tc_plan_end}\;
Build speculative queue: $\mathcal{Q}^{\mathrm{S}}_{i+1}\gets(\widehat{e}_r\mid\widehat{e}_r\notin\mathcal{R}_{i+1})$ in prediction order\nllabel{line:tc_spec_queue}\;
Run CPU experts while draining $\mathcal{Q}^{\mathrm{E}}_i$\nllabel{line:tc_exact_run}\;
When $\mathcal{Q}^{\mathrm{E}}_i=\varnothing$, run GPU experts and prefetch from $\mathcal{Q}^{\mathrm{S}}_{i+1}$ concurrently\nllabel{line:tc_spec_run}\;
At next exact routing, stop prefetching and clear $\mathcal{Q}^{\mathrm{S}}_{i+1}$; retain loaded experts\nllabel{line:tc_stop}\;
Update runtime rates $v^{\mathrm{C}},v^{\mathrm{G}},v^{\mathrm{T}}$\nllabel{line:tc_rates}\;
\end{algorithm}

\textbf{Speculative expert assignment.}
Speculative assignment uses the speculative ranker's ordered descriptor $\widehat{\mathcal{D}}_{i+1}^{(s)}$ to place predicted next-layer experts on GPU before their exact demand is known. The controller filters out experts already resident in the next layer's GPU expert pool and preserves the ranker's order of decreasing predicted token-load. Let $(\widetilde{e}_1,\ldots,\widetilde{e}_{\ell})$ denote the remaining candidates. The controller constructs the speculative transfer queue as: 
\begin{equation}
    \mathcal{Q}^{\mathrm{S}}_{i+1}
    =
    (\widetilde{e}_1,\ldots,\widetilde{e}_{\ell}).
    \label{eq:prefetch_transfer_queue}
\end{equation}
After all transfers in $\mathcal{Q}^{\mathrm{E}}_i$ complete, the controller transfers these candidates to GPU one by one while ongoing computation proceeds. Unlike exact assignment, which selects a transfer count from the current demand and runtime rates, speculative assignment follows the predicted priority for as long as the overlap window permits.

When the router of layer $i+1$ produces the exact routing result, the controller stops speculative transfers and clears the remaining entries in $\mathcal{Q}^{\mathrm{S}}_{i+1}$. Completed prefetches remain in the GPU pool and are included in the resident set used by the next exact assignment. The controller then prioritizes the new exact transfers and uses the prediction for layer $i+2$ to prepare the next speculative assignment.

\textbf{Overall runtime algorithm.}
Algorithm~\ref{alg:transfer_controller} summarizes this workflow. Lines~\ref{line:tc_prepare_start}--\ref{line:tc_prepare_end} identify and sort the current GPU misses, compute their accumulated loads $S_p$, and obtain the activated GPU-resident load $L_{\mathrm{R}}$. Lines~\ref{line:tc_search_start}--\ref{line:tc_search_end} scan all transfer counts $p\in\{0,\ldots,m\}$, estimate the CPU and GPU path latencies, and retain the count $p_i^*$ that minimizes their maximum. Lines~\ref{line:tc_plan_start}--\ref{line:tc_plan_end} convert this choice into the CPU-GPU execution plan and exact transfer queue, while line~\ref{line:tc_spec_queue} prepares the speculative queue from predicted experts.
Execution then follows the transfer priority: line~\ref{line:tc_exact_run} overlaps CPU execution with exact weight transfers, and line~\ref{line:tc_spec_run} starts GPU execution and speculative prefetching after those transfers complete. At the next exact routing event, line~\ref{line:tc_stop} stops speculative prefetching while retaining the experts already transferred. Finally, line~\ref{line:tc_rates} updates the runtime rates for subsequent assignment decisions.

\subsection{Inference Engine} \label{sec:pipelined_execution}

The inference engine controls the forward pass and realizes the CPU-GPU execution plan produced by the transfer controller. For each Transformer layer, it first performs the original normalization and attention computation on GPU. The attention output is then passed to the MoE block, where the router determines the selected experts and routing scores for every token. 
The router reports the exact activated experts and their loads to the transfer controller, which returns the execution plan $\Pi_i^{(s)}$.

After receiving the plan, the inference engine groups hidden states by their selected experts and dispatches each group to its assigned device. For every expert in $\mathcal{C}_i(p_i^*)$, the corresponding hidden states are sent to CPU and processed using the experts in host memory. The resulting outputs are then returned to GPU. In parallel, the controller transfers the missed experts in $\mathcal{G}_i(p_i^*)$ according to $\mathcal{Q}^{\mathrm{E}}_i$. Once the exact queue is empty, the inference engine executes these newly transferred experts on GPU. The controller then uses the available transfer window to prefetch next-layer experts while GPU computation proceeds.
Each selected expert produces an output $\mathbf{y}_{i,t,e}$ for every assigned token. After both executions complete and the CPU results return, the inference engine merges all expert outputs using the original routing scores:
\begin{equation}
    \mathbf{m}_{i,t}
    =
    \sum_{e\in\mathcal{T}_{i,t}}
    \alpha_{i,t,e}\mathbf{y}_{i,t,e},
    \qquad
    \mathbf{x}_{i+1,t}
    =
    \operatorname{Norm}\left(\mathbf{h}_{i,t}+\mathbf{m}_{i,t}\right).
    \label{eq:distributed_expert_merge}
\end{equation}
The resulting $\mathbf{X}_{i+1}$ becomes the input to the next Transformer layer, and the same workflow repeats. 
Besides, the inference engine reports the token-loads and the end-to-end CPU, GPU, and transfer times.
The transfer controller uses these measurements to update the online rates in Eq.~\eqref{eq:online_rate_update}, while the expert pool manager uses the token-load to maintain cache priority. The first Transformer layer performs exact assignment without an earlier transfer, and the final layer disables speculative assignments. 

\subsection{Expert Pool Manager} \label{sec:expert_pool_management}

The expert pool manager maintains the fixed-capacity expert pool in GPU memory. Before every assignment decision, it reports the latest resident set $\mathcal{R}_i$ to the transfer controller. Whenever the transfer controller brings in a new expert, the manager evicts an existing expert according to the eviction policy. 
The eviction policy aims to remove an expert that is unlikely to be activated again, thereby reducing the probability of reloading an expert shortly after its eviction. To capture both the recency and frequency of recent routing demand, the manager records an exponentially decayed token-load utility. At inference step $s$, it updates the utility as: 
\begin{equation}
    u_{i,e}^{(s)}
    =
    \gamma u_{i,e}^{(s-1)}
    +(1-\gamma)n_{i,e}^{(s)},
    \qquad 0\le\gamma<1,
    \label{eq:expert_cache_utility}
\end{equation}
where $n_{i,e}^{(s)}=0$ when expert $e$ is inactive. The utility emphasizes recent token-loads, which are the key signal for expert assignment. 
Before selecting an eviction victim, the manager excludes all experts that will be used by the current inference step according to the execution plan. Consequently, experts that repeatedly receive many tokens are more likely to remain in GPU memory, whereas experts with infrequent or low-load activations are replaced first.

\section{Performance Evaluation}\label{sec:evaluation}
\begin{table}[t]
    \centering
    \caption{MoE models used in the evaluation.}
    \label{tab:evaluation_models}

    \small
    \setlength{\tabcolsep}{3pt}
    \renewcommand{\arraystretch}{1.05}

    \begin{tabular}{lcccc}
        \toprule
        \multirow{2}{*}{Model}
        & \multicolumn{2}{c}{Parameters}
        & \multicolumn{2}{c}{Experts} \\
        \cmidrule(lr){2-3}
        \cmidrule(lr){4-5}
        & Total & Activated & Total & Activated \\
        \midrule
        Qwen3.5-35B-A3B     & 35B  & 3B  & 256 & 8  \\
        Qwen3-Next-80B-A3B  & 80B  & 3B  & 512 & 10 \\
        GLM-4.5-Air         & 106B & 12B & 128 & 8  \\
        \bottomrule
    \end{tabular}
\end{table}

\subsection{Experimental Setup} \label{sec:experimental_setup}

\textbf{Implementation.}
We implement \method on top of KTransformers~\cite{10.1145/3731569.3764843}, an efficient CPU-GPU collaborative inference framework. KTransformers keeps the complete expert weights in CPU host memory and retains only a subset of experts in GPU memory. It uses SGLang~\cite{3737916.3739916} for GPU execution and provides optimized MoE kernels for CPU expert computation. To support \method, we extend the framework with next-layer expert prediction and prefetching, online CPU-GPU expert assignment, and GPU expert-pool management.
We conduct all experiments on a server equipped with one NVIDIA RTX A6000 GPU with 48~GB of memory, one Intel Xeon Platinum 8358P CPU, and 512~GB of host memory. The CPU and GPU communicate through PCIe 4.0~$\times$16, which provides a theoretical peak bandwidth of 32~GB/s.

\textbf{Models and workloads.}
We evaluate \method using three representative MoE models: Qwen3.5-35B-A3B~\cite{qwen3.5}, Qwen3-Next-80B-A3B-Instruct~\cite{qwen2025next}, and GLM-4.5-Air~\cite{5team2025glm45agenticreasoningcoding}. As summarized in Table~\ref{tab:evaluation_models}, these models differ substantially in total parameter count, activated parameters, number of total experts, and number of activated experts. They therefore allow us to evaluate \method across diverse model scales and expert-activation patterns. All models use BF16 precision. None of them can be fully accommodated by the 48~GB GPU, making expert offloading necessary.
We construct the inference workloads from the validation split of MMLU~\cite{hendrycks2021measuringmassivemultitasklanguage} and vary the batch size to evaluate the systems under different serving loads. We use the disjoint training split only for the offline profiling such as collecting expert activation frequencies.

\textbf{Baselines.}
We compare \method against three representative expert-offloading approaches.

\textit{Mixtral-Offloading}~\cite{eliseev2023fastinferencemixtureofexpertslanguage} follows pure GPU execution and performs all expert computation on GPU. It predicts the experts required by the next layer from the current layer's intermediate representations and prefetches their weights to reduce exposed transfer latency.

\textit{Fiddler}~\cite{kamahori2025fiddlercpugpuorchestrationfast} represents static CPU-GPU collaborative execution. It profiles expert activation frequencies offline, places the most frequently activated experts on GPU, and executes every GPU-missed expert directly on CPU. Its expert placement remains unchanged during inference, and no experts are transferred at runtime.

\textit{Post-Fetch}~\cite{11420335} also executes every GPU-missed expert on CPU. After each inference iteration, however, it transfers the missed experts to GPU for possible reuse in subsequent decoding iterations.

To isolate the effects of their transfer and execution policies, we reproduce all baselines within the same KTransformers framework and assign them the same GPU expert-pool capacity as \method.

\textbf{Metrics.}
We quantify prefill performance using time to first token (TTFT) and prefill speed, and decode performance using time per output token (TPOT) and decode throughput~\cite{3691938.3691949,3691938.3691945}. TTFT measures the latency until the first output token is generated, while prefill speed measures the number of prompt tokens processed per second. TPOT measures the average time to generate each subsequent output token, while decode throughput measures the total number of output tokens generated per second across the batch. Lower TTFT and TPOT and higher prefill speed and decode throughput indicate better performance.

\begin{figure*}[t]
    \begingroup
    \setlength{\subfigcapskip}{2pt}
    \setlength{\subfigbottomskip}{0pt}
    \centering

    \makebox[\textwidth][c]{
    \includegraphics[height=0.12in]
        {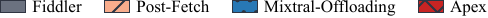}
    }

    \makebox[\textwidth][c]{
        \subfigure[Qwen3.5-35B-A3B\label{fig:prefill_qwen35}]{
            \includegraphics[width=0.31\textwidth]
                {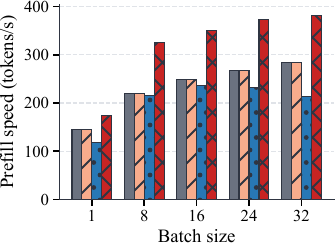}
        }\hspace{0.02\textwidth}
        \subfigure[Qwen3-Next-80B-A3B\label{fig:prefill_qwen_next}]{
            \includegraphics[width=0.31\textwidth]
                {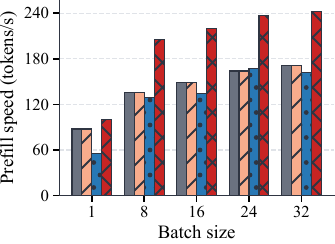}
        }\hspace{0.02\textwidth}
        \subfigure[GLM-4.5-Air\label{fig:prefill_glm45}]{
            \includegraphics[width=0.31\textwidth]
                {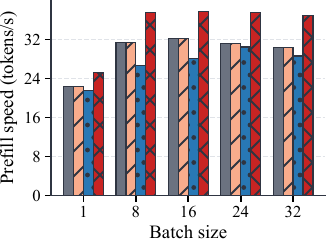}
        }
    }

    \caption{End-to-end prefill speed of \method and the baselines
    across batch sizes on three MoE models.}
    \label{fig:prefill_performance}
    \endgroup
\end{figure*}

\begin{figure*}[t]
    \begingroup
    \setlength{\subfigcapskip}{2pt}
    \setlength{\subfigbottomskip}{0pt}
    \centering

    \makebox[\textwidth][c]{
    \includegraphics[height=0.12in]
        {fig/fig1_throughput_legend.pdf}
    }

    \makebox[\textwidth][c]{
        \subfigure[Qwen3.5-35B-A3B\label{fig:decode_qwen35}]{
            \includegraphics[width=0.31\textwidth]
                {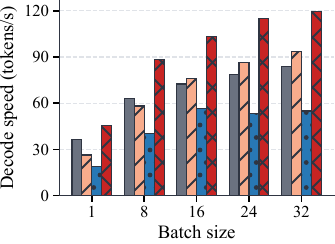}
        }\hspace{0.02\textwidth}
        \subfigure[Qwen3-Next-80B-A3B\label{fig:decode_qwen_next}]{
            \includegraphics[width=0.31\textwidth]
                {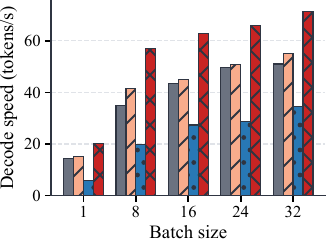}
        }\hspace{0.02\textwidth}
        \subfigure[GLM-4.5-Air\label{fig:decode_glm45}]{
            \includegraphics[width=0.31\textwidth]
                {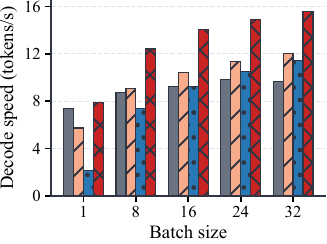}
        }
    }

    \caption{End-to-end decode throughput of \method and the baselines
    across batch sizes on three MoE models.}
    \label{fig:decode_performance}
    \endgroup
\end{figure*}

\subsection{Overall Performance} \label{sec:overall_performance}

We first compare the end-to-end performance of \method and the baselines across different batch sizes. We evaluate batch sizes of 1, 8, 16, 24, and 32 and report both prefill speed and decode throughput for every model and method.
For each model, we select a fixed GPU expert-pool capacity that fits within the available memory and use the same capacity for all methods. Specifically, the GPU pool accommodates 64 experts for Qwen3.5-35B-A3B, 56 experts for Qwen3-Next-80B-A3B-Instruct, and 12 experts for GLM-4.5-Air. The remaining GPU memory is reserved for non-expert model parameters, the KV cache, and runtime buffers. 
As shown in Figure~\ref{fig:prefill_performance}, \method consistently achieves the highest throughput across all models and batch sizes during the prefill stage.
Compared with the strongest baseline, it improves prefill speed by 31.2\% on average and by up to 51.9\%. Post-Fetch has the same prefill speed as Fiddler because its transfers occur only after an inference iteration and therefore cannot accelerate the current prefill stage. Mixtral-Offloading is generally slower than the two CPU-GPU collaborative baselines. This result shows that, under limited PCIe bandwidth, transferring every GPU-missed expert costs more than directly computing missed experts on CPU.
The prefill gain is relatively small at batch size 1, where few tokens are routed and the loads of activated experts provide limited opportunities for selective assignment. As the batch size increases, more experts are activated and their token-loads become increasingly uneven. \method exploits this diversity by transferring high-load misses to GPU, reducing CPU computation without introducing the excessive traffic of pure GPU execution.

Figure~\ref{fig:decode_performance} illustrates gains in the decode stage, in which \method improves throughput over the strongest baseline by 31.1\% on average and by up to 39.5\%.
Unlike in prefill, Post-Fetch can benefit from expert reuse across decoding iterations and therefore generally outperforms Fiddler. However, its inter-step transfer reflects the previous iteration and does not account for the current iteration. By combining speculative and exact expert assignment, \method avoids both limitations and remains the best-performing method.
The improvement also depends on the specific MoE architecture. Averaged over both phases and all batch sizes, \method improves over the strongest baseline by 36.4\% on Qwen3-Next, 34.1\% on Qwen3.5, and 22.9\% on GLM-4.5-Air. Qwen3-Next contains 512 experts and activates only 10 of them per token, providing fine-grained choices for moving high-load experts. GLM-4.5-Air has fewer, larger experts and activates 12B parameters per token. Each expert transfer is therefore more expensive and the assignment choice is coarser, leaving less room for improvement.

\begin{table*}[t]
    \centering
    \caption{Throughput under different GPU expert pool capacities at batch size 16.}
    \label{tab:memory_budget}
    \small
    \setlength{\tabcolsep}{2.6pt}
    \renewcommand{\arraystretch}{1.08}
    \begin{tabular*}{\textwidth}{@{\extracolsep{\fill}}llrrrrrrrrrr@{}}
        \hline
        & & \multicolumn{5}{c}{Prefill speed (tokens/s)} & \multicolumn{5}{c}{Decode throughput (tokens/s)} \\
        \cline{3-7}\cline{8-12}
        Model ($G_0$) & Method & $0.5G_0$ & $0.75G_0$ & $G_0$ & $1.25G_0$ & $1.5G_0$
        & $0.5G_0$ & $0.75G_0$ & $G_0$ & $1.25G_0$ & $1.5G_0$ \\
        \hline
        Qwen3.5 (64) & Fiddler & 177.60 & 205.90 & 249.60 & 316.50 & 383.80 & 56.10 & 65.10 & 72.44 & 83.80 & 93.90 \\
        & Post-Fetch & 177.60 & 205.90 & 249.60 & 316.50 & 383.80 & 59.19 & 69.31 & 75.85 & 94.33 & 104.33 \\
        & Mixtral-Off. & 181.94 & 201.69 & 236.20 & 258.71 & 284.24 & 46.14 & 52.88 & 56.51 & 60.43 & 63.40 \\
        & \method & \textbf{231.20} & \textbf{288.20} & \textbf{350.30} & \textbf{451.10} & \textbf{520.30} & \textbf{83.60} & \textbf{97.20} & \textbf{103.35} & \textbf{132.90} & \textbf{148.60} \\
        \hline
        Qwen3-Next (56) & Fiddler & 124.60 & 133.00 & 149.02 & 172.57 & 194.89 & 38.20 & 40.80 & 43.54 & 48.77 & 51.76 \\
        & Post-Fetch & 124.60 & 133.00 & 149.02 & 172.57 & 194.89 & 37.40 & 41.33 & 45.09 & 52.81 & 61.36 \\
        & Mixtral-Off. & 112.07 & 120.73 & 133.82 & 158.76 & 172.35 & 22.42 & 26.17 & 27.39 & 29.82 & 33.78 \\
        & \method & \textbf{165.10} & \textbf{191.40} & \textbf{219.98} & \textbf{258.75} & \textbf{289.97} & \textbf{56.60} & \textbf{61.30} & \textbf{62.88} & \textbf{77.04} & \textbf{84.44} \\
        \hline
        GLM-4.5-Air (12) & Fiddler & 29.37 & 30.83 & 32.09 & 32.47 & 33.49 & 8.59 & 8.91 & 9.25 & 9.50 & 9.66 \\
        & Post-Fetch & 29.37 & 30.83 & 32.09 & 32.47 & 33.49 & 9.82 & 10.18 & 10.44 & 11.06 & 12.36 \\
        & Mixtral-Off. & 23.74 & 26.88 & 28.06 & 30.29 & 32.94 & 7.79 & 8.46 & 9.22 & 10.68 & 11.66 \\
        & \method & \textbf{32.10} & \textbf{34.91} & \textbf{37.64} & \textbf{39.72} & \textbf{41.53} & \textbf{12.94} & \textbf{13.74} & \textbf{14.03} & \textbf{15.29} & \textbf{15.80} \\
        \hline
    \end{tabular*}
\end{table*}

\begin{figure}[t]
    \begingroup
    \setlength{\subfigcapskip}{2pt}
    \setlength{\subfigbottomskip}{0pt}
    \centering

    \makebox[\linewidth][c]{
        \includegraphics[height=0.12in]
            {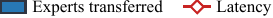}
    }

    \makebox[\linewidth][c]{
        \subfigure[Prefill stage\label{fig:transfer_latency_prefill}]{
            \includegraphics[width=0.475\linewidth]
                {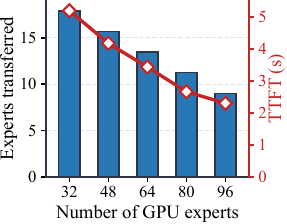}
        }\hspace{0.03\linewidth}
        \subfigure[Decode stage\label{fig:transfer_latency_decode}]{
            \includegraphics[width=0.475\linewidth]
                {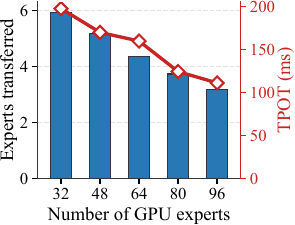}
        }
    }

    \caption{Impact of GPU-resident expert capacity on expert transfers
    and inference latency for Qwen3.5-35B-A3B.}
    \label{fig:transfer_latency}
    \endgroup
\end{figure}

\subsection{Performance across GPU Memory Budgets} \label{sec:memory_budget}

We next evaluate the effectiveness of \method under different GPU memory constraints. We fix the batch size to 16 and vary the capacity of the GPU expert pool. A smaller pool produces more GPU misses and places greater pressure on CPU execution and expert transfer, whereas a larger pool allows more activated experts to execute directly on GPU.
We use the capacity from Sec.~\ref{sec:overall_performance} as a practical, model-specific budget, denoted by $G_0$. Thus, $G_0$ is 64 experts for Qwen3.5, 56 for Qwen3-Next, and 12 for GLM-4.5-Air.
We sweep the pool capacity from $0.5G_0$ to $1.5G_0$ and conduct experiments, and the results are reported in Table~\ref{tab:memory_budget}.
As expected, increasing the resident expert budget improves every method by reducing GPU misses. More importantly, \method achieves the highest prefill and decode throughput at every budget for all three models. Relative to the strongest baseline in each setting, it improves prefill speed by 33.0\% and decode throughput by 39.2\% on average. The advantage persists from $0.5G_0$ to $1.5G_0$, showing that adaptive assignment is effective under both tight and relatively relaxed GPU memory budgets rather than being tuned to a single configuration.

To further understand the transfer behavior of \method, we record the number of transferred experts and the corresponding latency on Qwen3.5. We use TTFT for prefill and TPOT for decode in this analysis.
As shown in Figure~\ref{fig:transfer_latency}, increasing the GPU expert budget reduces both the number of transfers and inference latency. When the pool grows from 32 to 96 experts, the number of prefill transfers decreases from 17.9 to 9.0, while TTFT decreases from 5.19 to 2.30 seconds. During decode, the corresponding transfer count decreases from 5.94 to 3.18 and TPOT decreases from 197.6 to 111.1 milliseconds. 
With a larger resident pool, fewer activated experts miss on GPU. The CPU consequently receives a smaller workload, so \method needs to transfer fewer high-load experts to balance the concurrent CPU and GPU paths. The GPU-side MoE latency consists of exposed transfer followed by GPU expert computation. Since GPU expert computation is short relative to expert transfers, changes in the GPU path are dominated by the number of exposed transfers. The synchronized decrease in transfer count and latency therefore confirms that \method adapts its transfer volume to the available GPU budget.

\begin{table}[t]
    \centering
    \caption{Ablation results of \method on Qwen3.5-35B-A3B at batch size 16.}
    \label{tab:ablation}
    \small
    \setlength{\tabcolsep}{3.2pt}
    \renewcommand{\arraystretch}{1.08}
    \begin{tabular}{lrrrr}
        \hline
        & \multicolumn{3}{c}{Average layer latency (ms)} & Decode \\
        \cline{2-4}
        Configuration & CPU & GPU & Critical & tokens/s \\
        \hline
        No assignment & 3.074 & 0.295 & 3.074 & 72.44 \\
        SA only & 2.515 & 0.408 & 2.515 & 84.62 \\
        EA only & 2.220 & 2.320 & 2.320 & 90.31 \\
        Full \method & 2.130 & 1.930 & \textbf{2.130} & \textbf{103.35} \\
        \hline
    \end{tabular}
\end{table}

\subsection{Ablation Study} \label{sec:ablation}

We use Qwen3.5-35B-A3B to isolate the contributions of speculative assignment (SA) and exact assignment (EA) in \method.
We compare four configurations:
\textit{No assignment} disables both mechanisms and executes every GPU-missed expert on CPU.
\textit{SA only} predicts and transfers next-layer experts during the available overlap window, but does not make an exact assignment and won't make the GPU wait.
\textit{EA only} disables prediction and speculative prefetching but retains exact assignment. After exact routing, it selects GPU-missed experts for transfer according to the observed CPU, GPU, and transfer rates. 
\textit{Full \method} enables both mechanisms. SA transfers predicted high-load experts ahead of demand, while EA uses the exact routing result to determine the CPU-GPU assignment at runtime.

For each configuration, we report decode throughput and the average per-layer CPU and GPU path latency within the MoE module. The CPU path consists of CPU expert computation, whereas the GPU path includes GPU expert computation and exposed transfer latency. Since the two paths execute concurrently, we use the maximum of their average times to approximate the critical-path latency.
Table~\ref{tab:ablation} shows that SA and EA are complementary. With \textit{No assignment}, the GPU completes its resident-expert computation in only 0.30 ms, while the CPU path takes 3.07 ms and limits throughput to 72.44 tokens/s.
\textit{SA only} uses the overlap window to move predicted high-load experts to GPU without introducing GPU waiting. This reduces the CPU path to 2.52 ms and improves throughput by 16.8\%. However, the CPU path remains substantially longer than the 0.41 ms GPU path, indicating that speculative assignment cannot sufficiently balance the two paths.
\textit{EA only} achieves a closer balance by assigning experts according to actual demand and the observed execution rates. The CPU and GPU paths take 2.22 and 2.32 ms, respectively, and throughput improves by 24.7\% over \textit{No assignment}. However, all selected transfers begin after exact routing, exposing their latency on the GPU path and limiting room for improvement.
\textit{Full \method} combines the benefits of both mechanisms. SA overlaps expert transfers with preceding-layer computation, allowing more CPU work to move to GPU without extending the GPU path. EA then uses exact demand and GPU pool state to balance the remaining CPU computation against GPU transfer and computation. Together, they reduce the critical-path latency to 2.13 ms and achieve 103.35 tokens/s, outperforming \textit{No assignment}, \textit{SA only}, and \textit{EA only} by 42.7\%, 22.1\%, and 14.4\%, respectively.

\section{Related Work}\label{sec:related}
The large expert memory footprint of MoE models makes inference challenging under limited GPU memory. Expert offloading alleviates this constraint by storing expert weights in host memory and retaining only a subset on the GPU. Existing approaches primarily differ in how they execute activated experts that are absent from GPU memory.

\textbf{Pure GPU execution.}
Early frameworks, for example, DeepSpeed-Inference~\cite{10046087}, use host memory as a backing store and stage model parameters to GPU on demand. MoE-specific systems exploit sparse activation to transfer only the experts selected by the router. Mixtral-Offloading~\cite{eliseev2023fastinferencemixtureofexpertslanguage} combines expert caching with prefetching, which applies a future-layer router to an earlier hidden state and transfers predicted experts before their exact activation is known. MoE-Infinity~\cite{xue2025moeinfinityefficientmoeinference} instead records request-level expert-activation traces and uses the results to guide expert caching and prefetching.
Subsequent work improves the timeliness or accuracy of expert prediction. Pre-gated MoE~\cite{10.1109/ISCA59077.2024.00078} modifies the model so that each gating function selects experts for a later MoE block. ProMoE~\cite{song2025promoefastmoebasedllm} introduces a learned predictor and coordinates prediction, prefetching, and cache replacement to move likely experts into GPU memory before they are requested. fMoE~\cite{10.1145/3767295.3769319} further combines fine-grained expert-selection patterns with semantic information from the input to guide caching and prefetching. Despite these differences, expert computation remains on the GPU, and any GPU-missed expert must be transferred over PCIe before execution.

\textbf{CPU-GPU collaborative execution.}
Other systems exploit CPU computation to reduce expert weight transfers. Instead of transferring GPU-missed experts, they send the much smaller activations to CPU, execute the corresponding experts in host memory, and return their outputs to GPU. Fiddler~\cite{kamahori2025fiddlercpugpuorchestrationfast} demonstrates that this strategy can outperform weight transfers for memory-constrained, small-batch MoE inference. It profiles expert popularity to keep frequently activated experts resident on GPU. During inference, GPU-missed experts execute on CPU. KTransformers~\cite{10.1145/3731569.3764843} provides optimized heterogeneous kernels for executing selected model operators, including MoE experts, on CPU.
Recent work further refines expert residency for collaborative execution. Huang et al.~\cite{11420335}, referred to as \textit{Post-Fetch} in our evaluation, execute a GPU-missed expert on CPU for the current inference step and asynchronously transfer its weights into the GPU for potential reuse by subsequent inference steps. These approaches demonstrate that CPU computation can reduce costly expert transfers and enable both processors to contribute to MoE inference.

\vspace{-3mm}
\section{Conclusion}\label{sec:conclusion} 
This paper presents \method, a CPU-GPU collaborative inference system for memory-constrained MoE inference. \method uses the token-load of each expert and online measurements of CPU computation, GPU computation, and expert transfer to determine its execution device at runtime, avoiding both long CPU execution time and excessive transfer time.  
We implement \method on KTransformers and evaluate it on modern MoE models. Compared with the strongest baseline, Apex improves prefill speed by 31.2\% on average and up to 51.9\%, while increasing decode throughput by 31.1\% on average and up to 39.5\%. 
Its advantage persists across all evaluated GPU capacities. These results demonstrate that balancing expert transfer against CPU computation at runtime enables efficient inference of large MoE models under limited GPU-memory budgets.

\bibliographystyle{ACM-Reference-Format}
\bibliography{main}

\end{document}